\documentclass[10pt,twocolumn,letterpaper]{article}

\usepackage[pagenumbers]{cvpr} % To force page numbers, e.g. for an arXiv version

\usepackage{graphicx}
\usepackage{amsmath}
\usepackage{amssymb}
\usepackage{booktabs}

\usepackage[pagebackref,breaklinks,colorlinks]{hyperref}

\usepackage{comment}
\usepackage{multirow}
\usepackage{enumitem}
\usepackage{float}

\usepackage{array, makecell}

\usepackage[capitalize]{cleveref}
\crefname{section}{Sec.}{Secs.}
\Crefname{section}{Section}{Sections}
\Crefname{table}{Table}{Tables}
\crefname{table}{Tab.}{Tabs.}

\def\cvprPaperID{12159} % *** Enter the CVPR Paper ID here
\def\confName{CVPR}
\def\confYear{2022}

\begin{document}

%%%%%%%%% TITLE
\title{Young Labeled Faces in the Wild (YLFW): A Dataset for Children Faces Recognition}

\author{Iurii Medvedev$^*$\\
$^*$University of Coimbra\\
Institute of Systems \\
and Robotics\\
3030-194, Coimbra, Portugal\\
{\tt\small iurii.medvedev@isr.uc.pt}
% For a paper whose authors are all at the same institution,
% omit the following lines up until the closing ``}''.
% Additional authors and addresses can be added with ``\and'',
% just like the second author.
% To save space, use either the email address or home page, not both
\and
Farhad Shadmand$^*$\\
{\tt\small farhad.shadmand@isr.uc.pt}
\and
Nuno Gonçalves$^{*,\dagger}$\\
$^\dagger$Portuguese Mint and Official \\
Printing Office (INCM)\\
1000-042, Lisbon, Portugal\\
{\tt\small nunogon@deec.uc.pt}
}

\maketitle

%%%%%%%%% ABSTRACT
\begin{abstract}
    %face recognition - great advances
    Face recognition has achieved outstanding performance in the last decade with the development of deep learning techniques. 

    Nowadays, the challenges in face recognition are related to specific scenarios, for instance, the performance under diverse image quality, the robustness for aging and edge cases of person age (children and elders), distinguishing of related identities.

    %problems and bias
    In this set of problems, recognizing children's faces is one of the most sensitive and important. One of the reasons for this problem is the existing bias towards adults in existing face datasets. 
    
    %work explanation
    In this work, we present a benchmark dataset for children's face recognition, which is compiled similarly to the famous face recognition benchmarks LFW, CALFW, CPLFW, XQLFW and AgeDB.  
    We also present a development dataset (separated into train and test parts) for adapting face recognition models for face images of children.
    The proposed data is balanced for African, Asian, Caucasian, and Indian races. To the best of our knowledge, this is the first standartized data tool set for benchmarking and the largest collection for development for children's face recognition. Several face recognition experiments are presented to demonstrate the performance of the proposed data tool set.

\end{abstract}

%%%%%%%%% BODY TEXT
\section{Introduction}

Due to the significant progress of deep learning techniques in the last decade face biometrics has become one of the most accurate biometrics modalities. However, there are still many problems that address modern face recognition. For instance, the most challenging ones are the efficient distinguishing of relatives or twins, the impact of diverse image quality, and age bias.

%motivation 
Here the efficient recognition of children takes an important place. Its necessity is usually motivated by the problem of finding missing or abducted children and combating the children's exploitation. The children's involvement in criminal activity, both as a victim and as a perpetrator is also a problem, where face recognition can be useful. The recognition of newborns is important for protecting from swapping of newborns in hospitals and maternity homes as an alternative biometric solution to accompany currently existing techniques based on RFID technology.

%describing problems
%missing stats
For instance, the annual reports in many countries present a large number of missing children. United Kingdom police forces recorded 50k missing individuals in 2020/21 years \cite{missing_children_uk}. In the United States, an estimated 340k children are reported missing in 2021 \cite{missing_children_usa}. In the same year Government of Canada reported an estimated 28k children missing \cite{missing_children_canada}.

%finding the missing child
Face recognition technologies indeed already contribute to solving this problem. For instance recently in China a family, which spent more than three decades searching for their abducted son, has finally reunited with their child with the use of face recognition technologies \cite{missing_child_found_china}. A Chinese man who was kidnapped as a toddler was found after police synthesized his adult image based on his child photos and identified him in their database.

%children exploitation unicef
Another motivation to study face recognition for children's faces is children's exploitation. According to UNICEF in 2021 across 129 countries 4.4 million children had experienced violence (of the 2.3 million for whom disaggregated data are available, 53 percent are girls) with health, social work, or justice/law enforcement services. This number increased by 80\% compared to 2017. This significant growth indeed is also related to better availability of support but indeed it also discovers the large latent scale of this problem.

%crime with children
The more emergent problem is the children's involvement in criminal activity, both as a victim and as a perpetrator.
For instance, there is much evidence across the world about the involvement of children into drugs and the drug trade \cite{children_drugs, impact_drug_policies_children}.

%face recognition argentina
Here face recognition can already be successfully applied. For example, the live facial recognition system, which is deployed in Buenos Aires by the city government, utilizes the enrolled children identities \cite{fr_children_argentina}. Their database, known as the Consulta Nacional de Rebeldías y Capturas (National Register of Fugitives and Arrests), or CONARC includes known alleged offenders. The database is regularly updated with children's faces, where the age of the youngest alleged offender (who is cited for “crimes against persons (malicious)—serious injuries”) is around four years old.

%character of aging 
This evidence proves the importance of taking special measures for recognizing children's faces. Indeed the character of aging of children is different compared to adults. The maturation of children involves the nonlinear structure and shape changes of the skull and propositions of parts of a face. For instance, eyes grow rapidly right after birth. Then in several months their growth becomes more linear and undergoes an extra growth spurt during puberty \cite{maturation_children, maturation_children_2}. At the same time adults, aging has less extensive character and is usually limited to soft tissue changes like skin texture, hair color, and wrinkles appearance. 

%conclusion and bridge to our work
To sum up, children's facial images are different from adults ones, which is caused by physiological maturation process features. At the same time currently, many COTS (Commercial off-the-shelf) face recognition algorithms and respective academic advances are biased to recognize adults better than children \cite{NIST_FRVT}.
%there is still no such  tool in academia
In academia, a significant amount of face recognition benchmarks for adults exist \cite{LFW_dataset, CALFW, CACD_dataset, CPLFW, XQLFW, IJBA, IJBB, IJBC,agedb_30}. For the children's age group such testbeds (usually private) were also proposed \cite{Longitudinal_children, Auto_Face_Recognition_Children, Face_Recognition_In_Children, Ricanek_review}. However, the standard public tool is still not present in the academic community.

%introduce a dataset
In this work, we address the above problem and present a novel face data toolset, which is specifically focused on children's face recognition. The collected source data consists of wild children's face images of diverse ages, thus we choose the abbreviation YLFW (Young Labeled Faces in the Wild)\footnote{https://github.com/visteam-isr-uc/YLFW} for referencing our toolset and its components. Our toolset consists of two parts YLFW-Benchmark, and YLFW-Dev for different aspects of face recognition research for the children's age group.

%YLFW_Benchmark
To address the purpose of estimating the performance of face recognition algorithms against children's face images, we propose a YLFW-Benchmark dataset. The dataset consists of $\sim$ 10k images of $\sim$ 3k identities and it is accompanied by a 1-1 verification protocol, which includes 3k match and 3k non-match pairs.

%YLFW_dev
%existing datasets - include joint identities
To address the purpose of the development of face recognition systems, which are adapted to children's faces we propose the YLFW-Dev dataset which is split into YLFW-Dev-Train (2k identities, $\sim$ 76k images) and YLFW-Dev-Test ( $\sim$ 1k identities, $\sim$ 2k images) with disjoint identities.  
YLFW-Dev-Test is built similarly to YLFW-Benchmark but includes fewer images and comparison pairs. Its purpose is to support benchmarking in a case when YLFW-Dev-Train is used in the training data (YLFW-Benchmark should not be used in this case since it shares identities and images with YLFW-Dev-Train).

%comment about disjoint identities
To the best of our knowledge, our data toolset provides the first public benchmark that is directed at estimating the performance against the children's age group faces and the largest training dataset for the respective age group.
The important advantage of the proposed data is that it is not collected by the list of identities of celebrities.  That is why the face recognition algorithms, which are trained on famous academic datasets of celebrities (like CASIA-Webface\cite{casia_webface}, VGGFace2\cite{VGGface2}, MS-Celeb-1M\cite{ms_celeb_face}) can be tested on YLFW components following the correct open-set testing scenario (when images in the training and testing parts include images of disjoint identities). Standart benchmarks (like LFW\cite{LFW_dataset}, IJB\cite{IJBA,IJBB,IJBC} etc.) are usually based on the images of celebrities and can share the same identities with the training data. That is why we also argue that YLFW-Dev-Train indeed can be concatenated with the common academic face datasets for training face recognition networks.

%discuss features
Several features of the proposed data in YLFW should be noted. 

%race balanced
The proposed YLFW data is race balanced in different means. Namely, YLFW-Benchmark and YLFW-Dev-Test are race balanced by the number of pairs across the protocol. The original YLFW-Dev-Train is balanced only by the number of identities per race. That is why we also provide and employ the YLFW-Dev-Train-balanced, which is obtained by additional augmentation of images of fewer represented races in the original YLFW-Dev-Train.

%no gender balance
The gender differences between children are weaker and several works demonstrate that it is harder to estimate the gender of a child compared with an adult \cite{is_it_easy_to_rec_ch, classifying_adults_children}. For newborns, the gender property is usually hardly identified by the face image. That is why we do not prioritize gender balance in our work.

%No explicit longitudial data. the main purpose is not the cross aging
It is important to discuss the correlation of the proposed datasets to the cross-age recognition problem.  Due to the semi-automatic nature of data collection, our data toolset have only collateral age longitudinal property.
%in detail for benchmark
YLFW benchmark indeed contains pairs with the age difference, however, we do not explicitly control this effect since the main purpose of the dataset is estimation the performance biased to children's faces. Namely, the proposed benchmark is not explicitly focused on a cross-age recognition problem. 
% words that dev set contains cross-age labels
The YLFW-Dev set contains classes, where face images were collected with a large age gap between the sessions (and thus with a perceptible maturation of facial features), however, this is also an uncontrolled effect of the data collecting process. We do not combat this effect since it better approximates the introduced dataset to the real application scenarios.

\section{Related Work}

To introduce our methodology and results, we first need to discuss recent advances in face recognition, its benchmarking, and specificity when dealing with children's faces. 

\subsection{Face Recognition}

The ability to learn highly discriminative features from unconstrained images, which is provided by deep learning tools, facilitated the significant development of race recognition technologies in the last decade.

Convolutional neural networks (CNN) have become a standard tool for face recognition due to their high efficiency in solving pattern recognition problems \cite{ImageNet_cite}.  The training approach of such deep networks can vary, but the target is usually the same - to learn low-dimensional feature domain, where the sample discrimination may be performed with trivial similarity metrics. 

%metric learning approaches
Training usually proceeds on large face image datasets, where images are collected in wild conditions. Such datasets are usually scraped from the web. Some learning methods utilize the contrast between match/non-match pairs to learn the feature domain \cite{chopra_metric_paper, facenet}. These metric learning methods explicitly optimize the target similarity measure with no binding to particular class labels, however, they tend to require careful sample mining strategies to achieve robust convergence.

%classification approaches
However, most commonly deep learning face recognition is approached by solving the classification task on the identity-labeled training dataset. While optimizing the probabilities of class labels for training samples the required feature domain (carried by hidden layers) can also be learned. This is usually achieved by utilizing Softmax loss and its modifications for classification \cite{deepid_paper,deepid2_paper,deepid2_plus_paper}.

The performance of this technique can be significantly improved by increasing intra-class compactness and maximizing inter-class discrepancy from different perspectives. For example, by applying additional regularization for pushing intra-class features to their center \cite{centerface_paper}, or by introducing several kinds of marginal restrictions for inter-class variance  \cite{sphereface_paper, cosface_paper, arcface_paper, equalized_margin_paper}.

%sample specific
Modern approaches usually consider sample-specific strategies, which allow better control of a feature domain for achieving higher intra-class compactness and inter-class separation. 
For example, sample labeling may be performed by its hardness \cite{npcface, Huang_2020_CVPR}, additional data augmentation applied \cite{towards_face_recognition} or even by treating its deep features in a distributional manner (by specifying sample \textit{uncertainty}) \cite{probabilistic_embedding}.

%magface %Adaface %qualface
Quality-based loss function adaptations have been intensively studied in recent works.  Here the MagFace \cite{magface}, QualFace\cite{QualFace,qualface2} and AdaFace\cite{AdaFace} share conceptual similarities of the approaches. All these losses indeed modify the marginal-based softmax in a sample-specific way.  MagFace and AdaFace utilize the earlier reported correlation between the deep feature magnitude and sample quality \cite{addictive_margin_paper}.  AdaFace intends to approximate the image quality with feature norms. In contrast, MagFace learns a representation that aligns the feature norm with recognizability, which is closer in meaning to the sample difficulty than the image quality.
QualFace performs the loss adaptation with a more explicit and diverse set of quality metrics (including the generic image quality characteristics).

\subsection{Face Recognition for Children's Age Group}
The problem of bias in face recognition is diverse and includes many factors. Age bias is one of the most natural problems due to evident face feature changes during the process of maturation and further aging. 

%fgnet
The number of the pioneer works on age-invariant face recognition were performed using the FG-NET database \cite{fgnet_overview}, which is composed of a total of 1k images of 82 people with an age range from 0 to 69 and the largest age gap of 45 years. 
%cacd
Cross-Age Celebrity Dataset (CACD) \cite{CACD_dataset}, is a larger scale age-invariant dataset, which contains around 160k images of 2k celebrities with ages ranging from 16 to 62. 
However, the representation of children across this dataset is rather limited. 

%Table of datasets:

\begin{table*}[htbp]
\caption{Face datasets for children's face recognition}
\centering
\begin{tabular}{|l|c|c|c|c|c|c|c|}
\hline
\textbf{Dataset} & \makecell{ \textbf{Number of}\\ \textbf{Identities} } & \makecell{ \textbf{Number of}\\ \textbf{Images} } & \makecell{ \textbf{Age,}\\ \textbf{ years} } & \makecell{ \textbf{Acquisition} \\ \textbf{Type} }  & \textbf{Availability}& \makecell{ \textbf{Race}\\ \textbf{ Balanced} }& \makecell{ \textbf{Longi-}\\ \textbf{ tudinal} } \\ \hline
FG-NET \cite{fgnet_overview}     & 82 & 1002 & 6-18 & In the Wild &      Public   & No &   Yes   \\ \hline
AgeDB \cite{agedb_30}   &  568             &   16k                &  1 - 101  & In the Wild      & By Request  & No &   Yes  \\ \hline
YFA \cite{Face_Recognition_In_Children}  &   231   &    2293    &   3 - 14    & Controlled &   To be published  & No &   Yes \\ \hline
ITWCC \cite{Ricanek_review}      & 304 & 1.705 & 3+  & In the Wild  &    Private   & No &   Yes    \\ \hline
NITL \cite{Auto_Face_Recognition_Children}       & 314 & 3.144 & 3-5 & Controlled  &    Private & No &   Yes  \\ \hline
CLF \cite{Longitudinal_children} & 919 & 3.682 & 2-6 & Controlled &    Private & No &   Yes   \\ \hline 
LCFW \cite{Double_Blinded_Finder} & 6k & 60k & 1.5-9 & In the Wild &    Private  & No &   No  \\ \hline \hline
YLFW-Benchmark      & 3 069 & 9 810 & 0 - $\sim18$ & In the Wild &   Public  & Yes &   No  \\ \hline
YLFW-Dev-Train        & 2000 & 75k  & 0 - $\sim18$ & In the Wild &   Public  & Yes &   No   \\ \hline
YLFW-Dev-Test        & 1 016 & 1 887 & 0 - $\sim18$ & In the Wild &    Public & Yes &   No   \\ \hline
YLFW-Dev-Train-B.       & 2000 & 120k & 0 - $\sim18$ & In the Wild &   Public & Yes &   No    \\ \hline
\end{tabular}
\label{datasets-table}
\end{table*}

%bridge to children face images
There is a number of works that specifically focused on children's face recognition. The general trend in most of the works is also directed at performing a longitudinal study. Many works contributed with collected or web-scraped datasets of children's face collections. Usually, the provided datasets remain private. The list of currently existing datasets is presented in Table \ref{datasets-table}.

%NITL
In one of such works, Best-Rowden et al. proposed a NITL (Newborns, Infants, and Toddlers Longitudinal) face image dataset \cite{Auto_Face_Recognition_Children}, which was collected by the authors during several sessions. The database contains 314 subjects in total in the age range of 0 to 4 years old. The dataset is race biased to Indian faces. The experiments of this work proved the extreme complexity of recognizing newborns.

%Longitudinal_children
Another dataset Children Longitudinal Face (CLF) \cite{Longitudinal_children} was developed to cover a different children's age group for face recognition. CLF contains 3.5 k face images of 1k children in the age group of 2 to 18 years. Within this group, authors observed better performance than in \cite{Auto_Face_Recognition_Children} for the age group 0 to 4 years. Another interesting result is that girls in the CLF dataset have higher overall genuine scores and appear to be easier to recognize than boys.

%Face_Recognition_In_Children longitudinal
Bahmani and Schuckers also contributed to the Young Face Aging (YFA) dataset for analyzing the performance of face recognition systems over short age gaps in children \cite{Face_Recognition_In_Children}. The dataset is collected in the controlled acquisition and longitudinal time conditions and intended to be public via the BEAT research platform \cite{beat2017} (but not available at the moment of publication of this work). The authors demonstrated the positive correlation between face recognition performance decay and the age gap between the gallery and probe images in children, even at the short age gap of 6 months. At the same time, the authors did not observe a significant relationship between gender and match scores either in their dataset.

%Double_Blinded_Finder
In another work, Jin et al. developed a system for finding missing children without the exposure of photos on the web \cite{Double_Blinded_Finder}. To adopt the deep face representation for the system it was fine-tuned on the LCFW dataset collected by the authors. The LCFW dataset is collected by scraping the professional photo albums website and contains 60K images with 6K unique identities.

%ricanek review
Ricanek et al. proposed the In-the-Wild Child Celebrity (ITWCC) database, which is a collection of longitudinal wild face images of celebrities \cite{Ricanek_review}. It contains 304 subjects and 1705 images. The subject’s ages within this dataset range from 5 months to 32 years.  Also, the authors reviewed several face recognition algorithms on this dataset and showed that aging in non-adults is a challenging task for face recognition algorithms.

%comparizon to our work
In contrast to the above works, we do not explicitly focus on the longitudinal study and mainly consider the problem of age bias. From that perspective, we intend to propose a standardized benchmark for the testing and development of face recognition systems for children's faces.

\subsection{Face Recognition Benchmarks}
% Intro
Modern face recognition deep networks are trained on large labeled collections of face images and the resulting performance is usually estimated on separate datasets with disjoint identities.

% Scenarios
The most generic scenario for the benchmarking of face recognition systems is 1-1 verification. The respective benchmarks include the collection of face images with a pairing list (protocol), where each pair is given a match/non-match (by identity) label. In this work, we focus only on considering 1-1 verification. The number of existing benchmarks, which follows more sophisticated scenarios (like 1-N identification) is significantly less.  

% Public benckmarks
The most used benchmark data toolsets are usually freely distributed on the web.
%LFW 
One of the first and most popular face recognition benchmarks is Labeled Faces in the Wild (LFW), which is combined of 3k match and 3k non-match pairs\cite{LFW_dataset}.
The data in the LFW have natural variability of wild face image characteristics (like pose, lighting, focus, resolution, facial expression, age, gender, race, accessories, make-up, occlusions, and background). However, LFW does not cover all the aspects of the 1-1 face verification performance and also includes unwanted biases (for instance, the average age difference between the matched and non-matched pairs).

That is why several revisited variants appeared. They usually tend to enhance the hardness of correct verification (both inter-class and intra-class), which results in a significant performance decrease.
%CALFW
For instance, CALFW (Cross-Age Labeled Faces in the Wild) \cite{CALFW} is designed to reduce the age difference between match and non-match pairs and test the robustness of face recognition algorithms to face aging.

%CPLFW
The face pose difference is emphasized in CPLFW (Cross-Pose Labeled Faces in the Wild).
This dataset follows the LFW and provides a more realistic consideration of pose intra-class variation and fosters the research on cross-pose face verification in an unconstrained situation. 

%CFP-dataset
The problem of pose variation can be also studied with CFP dataset (Celebrities in Frontal-Profile) \cite{cfp-paper}. This dataset approaches the general pose variation problem by considering its extreme cases.  Images in CFP are matched to specific \textit{frontal} and \textit{profile} poses.   

%XQLFW
With the recently increased interest in image quality in face recognition, the XQLFW dataset (Cross-Quality Labeled Faces in the Wild) appeared \cite{XQLFW}. In contrast to increasing the variance of generic face image characteristics (pose, age)  XQLFW maximizes the quality difference between the samples in the protocol pairs. The quality variance is achieved by synthetic image degradation. Such an approach indeed can lead to a decrease of similarity in match pairs and intra-class similarity in general. This dataset helps to test the robustness of face recognition systems against image quality.

% Semipublic benckmarks
Some benchmarks are distributed by the institutional request with a license agreement. 
%RFW
For instance, Racial Faces in-the-Wild (RFW) \cite{RFW_Dataset} is a testing database for studying racial bias in face recognition. It provides for four testing race subsets, namely Caucasian, Asian, Indian, and African.  Each subset contains about 3000 individuals with 6000 image pairs in protocols for face verification. 

% AgeDB
AgeDB dataset \cite{agedb_30} was collected to investigate the aging problem in face recognition. It also provides protocols with pairs that include images with different age gaps to test the robustness of face recognition algorithms against aging.

% IJB
IJB set of benchmarks (IJB-A \cite{IJBA}, IJB-B \cite{IJBB}, IJB-C \cite{IJBC}) provide a large-scale test (both by the number of images and the length of pair list) on several face recognition tasks, challenges, and scenarios. Face images are collected with wide variations in pose, illumination, expression, resolution, and occlusion. 
However, the large size of these sets of benchmarks may become a problem from the perspective of timing and computing resources. For instance, the 1-1 verification protocol in IJB-C requires extracting 500k face embeddings and then performing 15M comparisons. At the same time, it is imbalanced in terms of the amount of match/non-match pairs.

%private benchmarks
Some fully private benchmarks do not distribute data but provide a platform for submitting face recognition algorithms. The data processing and the performance estimation are proceeded on the facilities of a benchmark provider. For instance, NIST FRVT \cite{NIST_FRVT} is a large-scale benchmark that includes protocols for both 1-1 verification and 1-N identification.  The results of these benchmarks report many aspects of the real face recognition performance for submitted algorithms. According to the NIST FRVT reports many submitted algorithms are biased to recognize adults better than children.

%conclusion
In our work we propose a compact toolset (similar to LFW by size) for investigating and combating the negative effect of such age bias. We approach the above problem first, by creating a benchmark for children's face recognition, which can become important for estimating the respective bias in face recognition systems. Second, we propose a novel data tool set for the development of face recognition approaches for children's faces.

\section{Methodology} 
\label{Methodology}

In this section, we discuss the methodology of data collection, combining benchmark protocols and the result datasets of this work. 

%We are not disclosing the details of collecting the identity list however its unwanted feature 

%In this section the word \textit{"model"} is used in a connotation of a term \textit{"photomodel"} - a person which is a subject of photography.

\subsection{Data collecting} 
\label{sec:data_collecting}
%The data collecting for YLFW started with the assembling of the identity list. 
The raw data acquisition follows the generic pipeline of collecting face datasets in computer vision. The face images are web scraped by iterating the list of identity references. In the most straightforward scenario such a list of identities is combined with the names of celebrities \cite{casia_webface, VGGface2}. In our work instead of proceeding with the collection by the list of known celebrities, the identities references are web scraped by a specific set of keywords.  These identity references are anonymous and for each one, a noisy set of images is downloaded.  
%filtering labels by clustering
The raw data is then filtered by hierarchical clustering to extract the main cluster for each identity label.

%race balance
For achieving a race balance the data collection proceeds with a race separation of the requests. We follow Wang et al. \cite{RFW_Dataset} in the definition of the race list and consider the following ones: African, Asian, Caucasian, and Indian. In our collecting process, we also observe that available data diversity for races decreases in the following order: Caucasian, Asian, African, and Indian.

%race observation
Indeed the real race separation is not discrete but smooth across the globe. In our data collecting process, we observe that for the selected list of races the most sensitive gap is between the Asian and Indian races. This observation may be caused by geographical closeness of Asian and Indian ethnic groups and also pitfalls of racial categories definition \cite{FR_race_categories}. 
%TODO find some race classification work with similar conclusion

\begin{figure}[htbp]
\centering
  \mbox{\subfloat[]{
            \label{fig:lfw-samples}
            \includegraphics[width=0.35\linewidth]{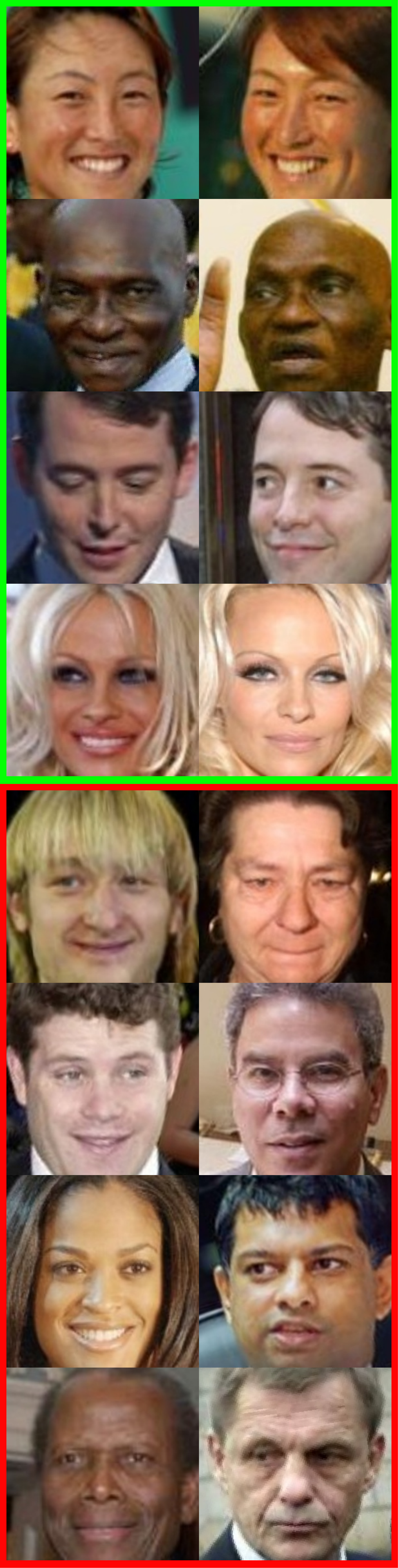}}}
  \mbox{\subfloat[]{
            \label{fig:ylfw-samples} 
            \includegraphics[width=0.35\linewidth]{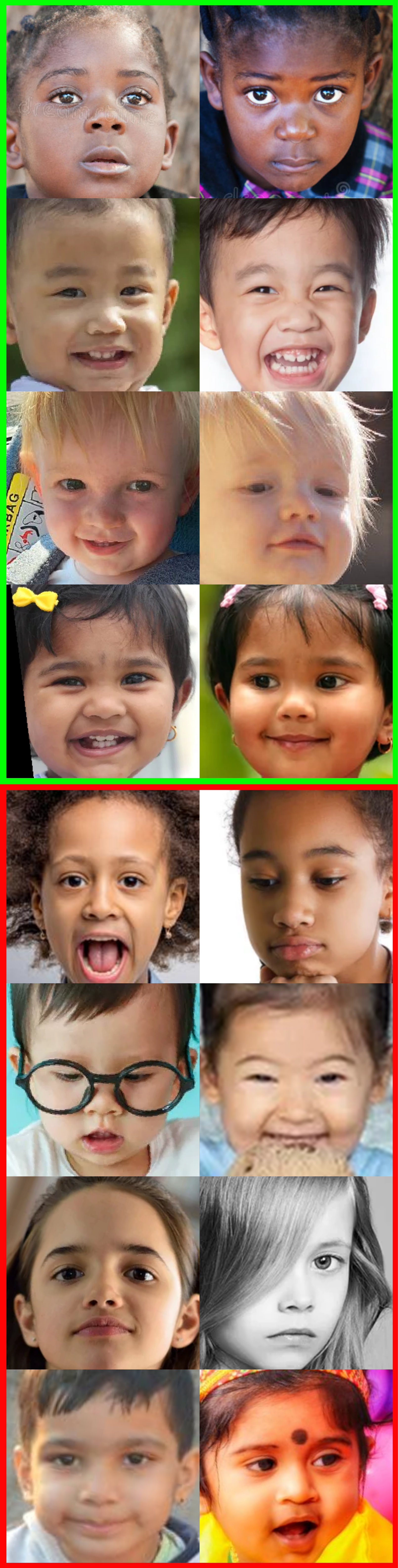}}}
%\vspace*{2pt}
\caption{Examples of pairs in LFW(a) and YLFW-Benchmark(b). Match pairs are in the green rectangle. Non-match pairs are in the red rectangle}
\label{fig:lfw-ylfw-samples}
\end{figure}

\subsection{Pairing methodology} 

In order to construct the 1-1 verification benchmark the image pair list (protocol) should be defined. In our work, we follow the semiautomatic process, where the proposed pairs are selected randomly. However, the proposed pairs are exposed to a human user verifier, who can accept or reject the pair. The process is repeated until the required number of approved pairs is achieved. All pairs indeed are also manually controlled with the intention to avoid extreme cases of bad quality, extremely easy match pairs, and rare but possible cases of mislabeling in the result data. %This process is not equal to human performance since the cross-label indication is available to the person, who performs the verification. 

The exact selection algorithm is slightly different for match and non-match pairs. In the case of match pairs first, the list of identities is combined. Then a single identity is randomly picked. Then two different images are randomly selected from this identity and exposed to the human user. The user selects one of three options: "accept"; "accept and remove"; "reject".  In the case of the "accept" decision, the image pair is appended to the protocol. In the case of the "accept and remove" decision, the respective identity is also removed from the identity list. The "reject" decision results in simple skipping of the proposed pair. After proceeding with the decision the cycle repeats.   

In the case of collecting non-match pairs, two equal lists of identities are generated. Then from each list, a random identity is picked. If the selected identities are not equal and this specific combination of identities is not present in the protocol, then for both identities a random image is then selected to be exposed to the human user as a proposed pair. The user selects from the same decision list as for match-pairs, however here in the case of the "accept and remove" decision the selected identities are removed from their respective lists.  After proceeding with the decision the cycle repeats.  

The pairing procedure was made by the first author for the protocols presented in this work.

\begin{figure*}[htbp]
\centering
  \mbox{\subfloat[]{
            \label{fig:benchs-benchs-1}
            \includegraphics[width=0.315\linewidth]{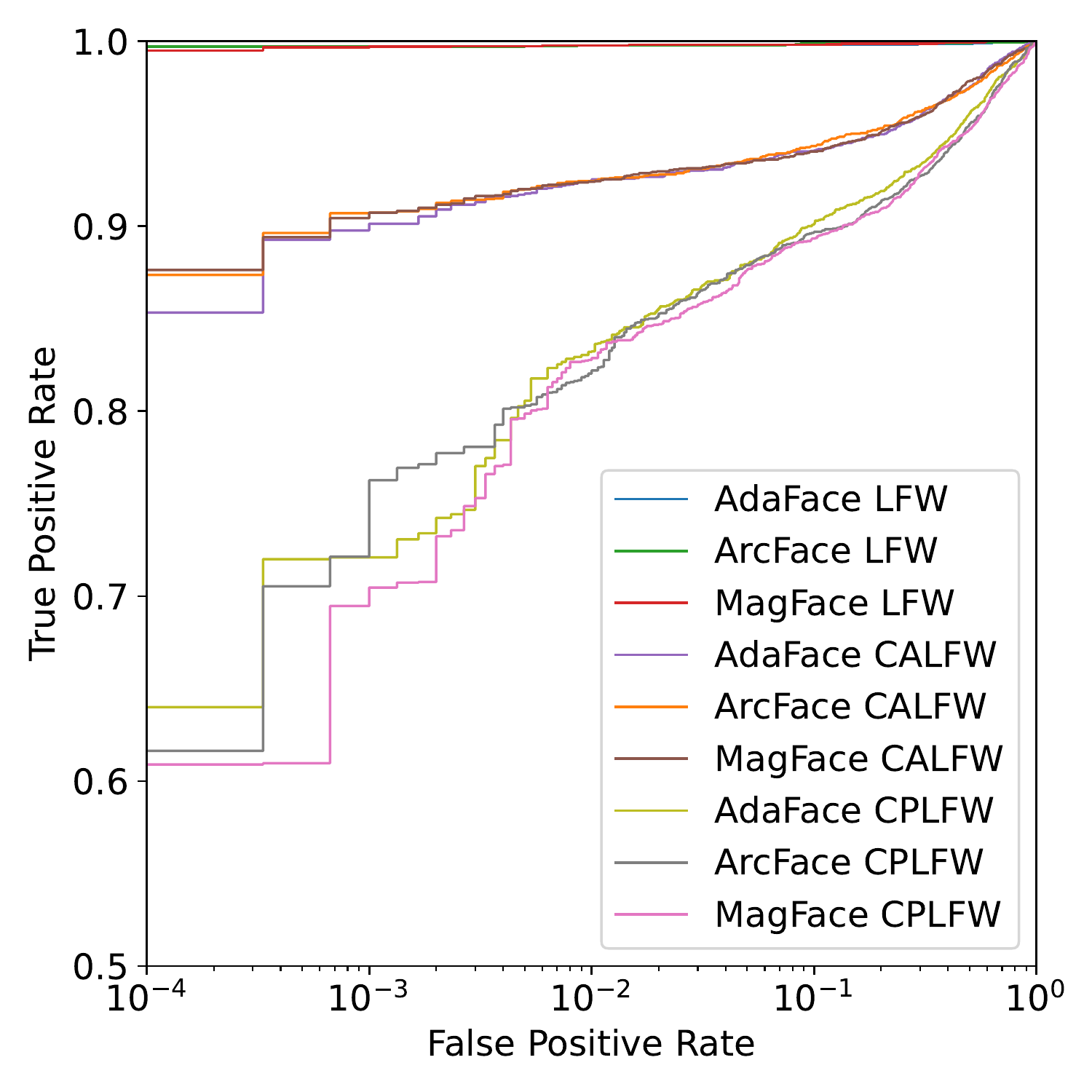}}}
  \mbox{\subfloat[]{
            \label{fig:benchs-benchs-2} 
            \includegraphics[width=0.31\linewidth]{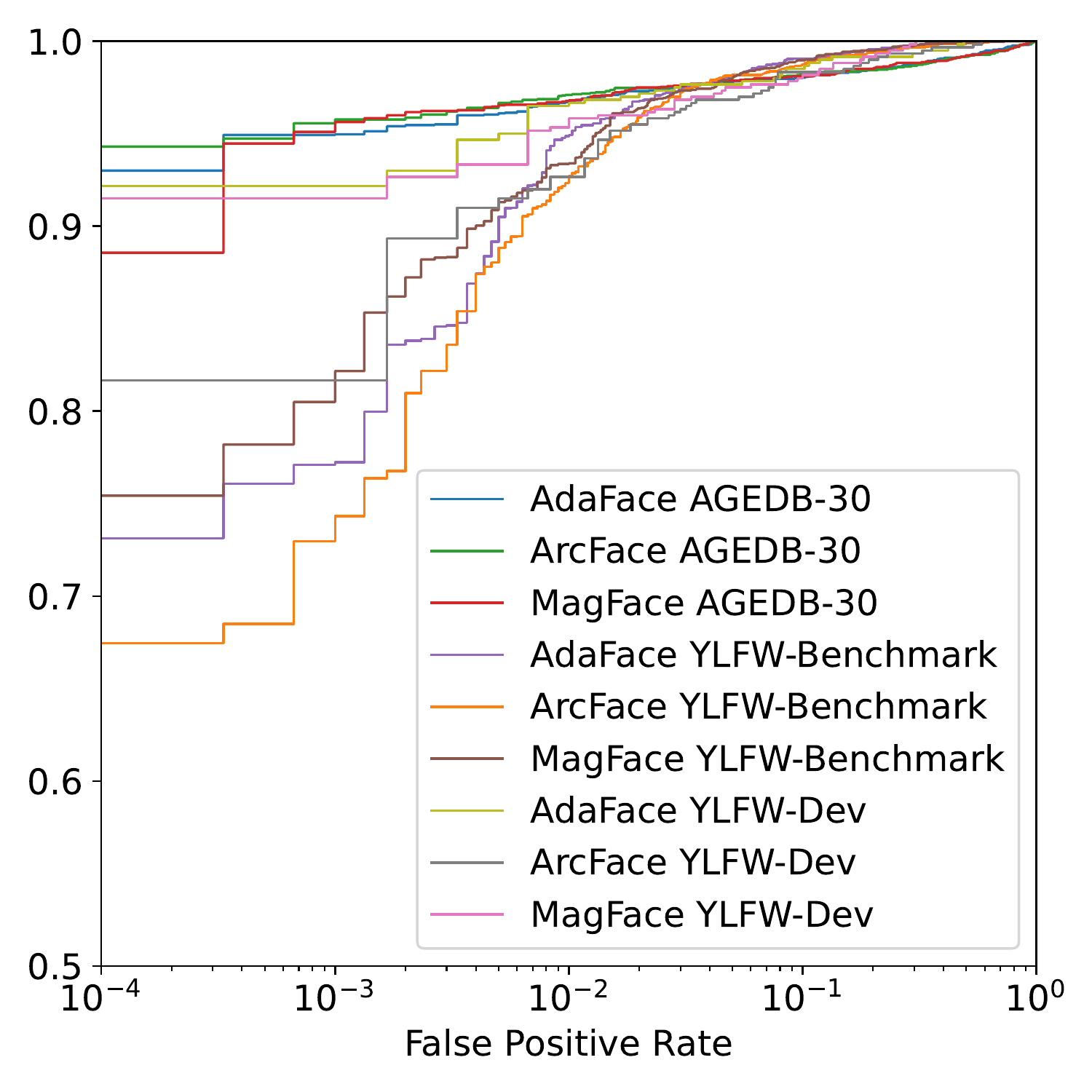}}}
  \mbox{\subfloat[]{
            \label{fig:benchs-benchs-3} 
            \includegraphics[width=0.31\linewidth]{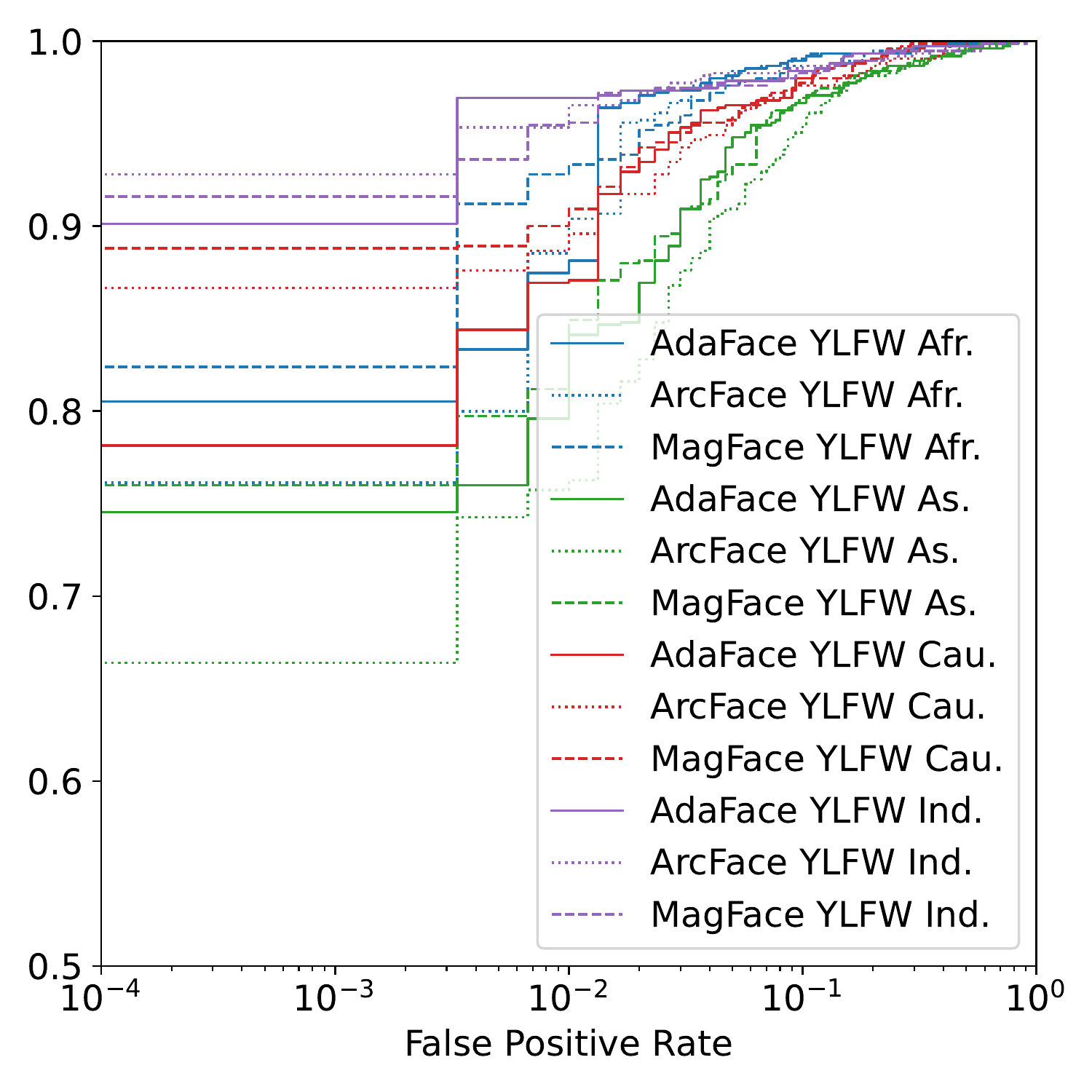}}}
\vspace*{2pt}
\caption{ROC curves of ResNet-50 network trained no MS1MV2 with different loss functions (AdaFace, ArcFace, MagFace). a) - (LFW, CALFW, CPLFW), b)- (AGEDB-30, YLFW-Benchmark, YLFW-Dev-Test) and c) - (race separated parts of YLFW-Benchmark, where Afr. - African, As. - Asian, Cau. - Caucasian, Ind. - Indian).  }
\label{fig:benchs-bench}
\end{figure*}

% Please add the following required packages to your document preamble:
% \usepackage{multirow}
\begin{table*}[htbp]
\centering
\caption{Performance metrics for LFW, CALFW, CPLFW, AGEDB-30, YLFW-Benchmark and YLFW-Dev-Test of ResNet-50 network trained no MS1MV2 with different loss functions (AdaFace, ArcFace, MagFace).}
\vspace*{2pt}
\setlength\tabcolsep{1.5pt}
\begin{tabular}{|c|cccccccccccc|}
\hline
\multirow{3}{*}{\makecell{ \textbf{Train}\\ \textbf{Dataset} }} & \multicolumn{12}{c|}{FNMR@FMR = $\alpha$, EER, AUC of ROC}                                                                                                                                                                                                                                                   \\ \cline{2-13} 
                  & \multicolumn{4}{c|}{LFW}                                                                         & \multicolumn{4}{c|}{CALFW}                                                                         & \multicolumn{4}{c|}{CPLFW}                                                    \\ \cline{2-13} 
                  & \multicolumn{1}{c|}{$\alpha$=$10^{-1}$} & \multicolumn{1}{c|}{$\alpha$=$10^{-2}$} & \multicolumn{1}{c|}{EER} & \multicolumn{1}{c|}{AUC} & \multicolumn{1}{c|}{$\alpha$=$10^{-1}$} & \multicolumn{1}{c|}{$\alpha$=$10^{-2}$} & \multicolumn{1}{c|}{EER} & \multicolumn{1}{c|}{AUC} & \multicolumn{1}{c|}{$\alpha$=$10^{-1}$} & \multicolumn{1}{c|}{$\alpha$=$10^{-2}$} & \multicolumn{1}{c|}{EER} & AUC \\ \hline
                  
                  AdaFace & \multicolumn{1}{c|}{0.0020} & \multicolumn{1}{c|}{0.0023} & \multicolumn{1}{c|}{0.0026} & \multicolumn{1}{c|}{0.9988} & \multicolumn{1}{c|}{0.0590} & \multicolumn{1}{c|}{0.0746} & \multicolumn{1}{c|}{0.0633} & \multicolumn{1}{c|}{0.9725} & \multicolumn{1}{c|}{0.0989} & \multicolumn{1}{c|}{0.1676} & \multicolumn{1}{c|}{0.0989} &  0.9515\\ \hline
                  ArcFace& \multicolumn{1}{c|}{0.0013} & \multicolumn{1}{c|}{0.0023} & \multicolumn{1}{c|}{0.0030} & \multicolumn{1}{c|}{0.9990} & \multicolumn{1}{c|}{0.0566} & \multicolumn{1}{c|}{0.0756} & \multicolumn{1}{c|}{0.0616} & \multicolumn{1}{c|}{0.9722} & \multicolumn{1}{c|}{0.1029} & \multicolumn{1}{c|}{0.1780} & \multicolumn{1}{c|}{0.1029} & 0.9478 \\ \hline
                  
                  MagFace& \multicolumn{1}{c|}{0.0020} & \multicolumn{1}{c|}{0.0023} & \multicolumn{1}{c|}{0.0026} & \multicolumn{1}{c|}{0.9992} & \multicolumn{1}{c|}{0.0596} & \multicolumn{1}{c|}{0.0759} & \multicolumn{1}{c|}{0.0639} & \multicolumn{1}{c|}{0.9724} & \multicolumn{1}{c|}{0.1066} & \multicolumn{1}{c|}{0.1713} & \multicolumn{1}{c|}{0.1049} & 0.9460 \\ \hline \hline

                & \multicolumn{4}{c|}{AGEDB-30}                                                                         & \multicolumn{4}{c|}{YLFW-Benchmark}                                                                         & \multicolumn{4}{c|}{YLFW-Dev-Test}                                                     \\ \hline
                  AdaFace & \multicolumn{1}{c|}{0.0196} & \multicolumn{1}{c|}{0.0320} & \multicolumn{1}{c|}{0.0256} & \multicolumn{1}{c|}{0.9906} & \multicolumn{1}{c|}{0.0093} & \multicolumn{1}{c|}{0.0506} & \multicolumn{1}{c|}{0.0276} & \multicolumn{1}{c|}{0.9955} & \multicolumn{1}{c|}{0.0150} & \multicolumn{1}{c|}{0.0333} & \multicolumn{1}{c|}{0.0266} & 0.9948 \\ \hline
                  
                  ArcFace & \multicolumn{1}{c|}{0.0176} & \multicolumn{1}{c|}{0.0290} & \multicolumn{1}{c|}{0.0250} & \multicolumn{1}{c|}{0.9903} & \multicolumn{1}{c|}{0.0126} & \multicolumn{1}{c|}{0.0726} & \multicolumn{1}{c|}{0.0300} & \multicolumn{1}{c|}{0.9945} & \multicolumn{1}{c|}{0.0166} & \multicolumn{1}{c|}{0.0733} & \multicolumn{1}{c|}{0.0333} & 0.9929\\ \hline
 
                  MagFace& \multicolumn{1}{c|}{0.0190} & \multicolumn{1}{c|}{0.0320} & \multicolumn{1}{c|}{0.0246} & \multicolumn{1}{c|}{0.9906} & \multicolumn{1}{c|}{0.0103} & \multicolumn{1}{c|}{0.0659} & \multicolumn{1}{c|}{0.0276} & \multicolumn{1}{c|}{0.9956} & \multicolumn{1}{c|}{0.0183} & \multicolumn{1}{c|}{0.0416} & \multicolumn{1}{c|}{0.0316} & 0.9950 \\ \hline

\end{tabular}
\label{performance_table_bench}
\end{table*}

\subsection{YLFW-Benchmark database} 

Following the above pairing procedure for the full collected data, we assemble the YLFW-Benchmark database.  To control the race balance this process of pairing is performed separately for different races and their cross combinations (for non-match pairs).
Namely, the resulting protocol includes 750 match pairs for each of the selected races and 300 non-match pairs for each cross-race combination.   
In total, the resulting protocol includes 3000 match and 3000 non-match pairs, which are based on 9810 images of 3 069 identities. 
Some examples of the collected pairs are presented in Fig. \ref{fig:lfw-ylfw-samples}.

% %race separated TODO
% In order to investigate the data differences between races in our experiments we also consider the subset benchmarks separated by race: YLFW-Benchmark-African, YLFW-Benchmark-Asian, YLFW-Benchmark-Caucasian, YLFW-Benchmark-Indian. Each of these tests include 750 match pairs and 300 non match pairs.

%Table of datasets:

\begin{table}[!htbp]
\caption{Summafy of the proposed dataset}
\centering
\begin{tabular}{|c|c|c|c|c|}
\hline
\makecell{ \textbf{YLFW}\\ \textbf{component} } & \makecell{ \textbf{Number of}\\ \textbf{Identities} } & \makecell{ \textbf{Number of}\\ \textbf{Images} } & \textbf{Purpose}  \\ \hline

Benchmark      & 3 069 & 9 810 & Testing  \\ \hline
Dev-Train        & 2000 & 75k  & Training    \\ \hline
Dev-Test        & 1 016 & 1 887  & Testing   \\ \hline
\makecell{ Dev-Train- \\ Balanced}       & 2000 & 120k & Training      \\ \hline
\end{tabular}
\label{ylfw-datasets-table}
\end{table}

\subsection{YLFW-Dev database} 

The goal of the development YLFW-Dev database is to provide separate parts for training and testing. The efficient training of deep networks for face recognition usually requires a large number of images per class (in the ideal scenario 10-50 images per identity \cite{celeb500k}). 

That is why basing on the collected data we select identities with the largest "samples per class" value to collect 500 identities per race. Thus the resulting YLFW-Dev-Train dataset is then only balanced by the "identities per race" parameter. That is we also design YLFW-Dev-Train-Balanced, which is obtained by random  additional augmentations of images of fewer represented races in the original YLFW-Dev-Train. The commonly used types of augmentations were employed to obtain YLFW-Dev-Train-Balanced (horizontal flip, brightness and contrast control, slight image rotation, noise injection).

The remaining part ("tail") of our collected data is used to combine a small benchmark with identities disjoint from the training part. The resulting YLFW-Dev-Test is then assembled similarly to YLFW-Benchmark and includes 1887 images of 1016 identities. Similarly to YLFW-Benchmark the resulting protocol is race balanced and contains 150 match pairs per race and 60 non-match pairs per each cross-race combination (1200 pairs in total).

\section{Experiments and Results}

We have performed several experiments with our data toolset. First, we evaluated several recent face recognition models on YLFW-Benchmark. Next, we trained several face recognition models on popular academic face recognition datasets, and their copies concatenated with YLFW-Dev-Train to demonstrate the benefits of such adaptation to the recognition of children's face images.

%following ISO we report
We report the performance by FNMR@FMR = $\alpha$ and also include additional metrics such as the Equal Error Rate (EER) of Detection Error Trade-off (DET) and Area Under Curve (AUC) of Receiver Operating Characteristic (ROC).

%list benchmarks
We estimate the performance on our generated tests YLFW-Benchmark and YLFW-Dev-Test in a combination with several similar compact face recognition benchmarks: LFW, CALFW, CPLFW, and ADEDB-30.

%comment regarding YLFW dev
The experiments on both YLFW-Benchmark and YLFW-Dev-Test in Section \ref{exp-YLFW-Benchmark} are performed indeed to compare those two tests and demonstrate compromises which are made in the development of YLFW-Dev-Test, compared to its larger companion.

At the same time in Section \ref{exp-YLFW-Dev}, the identities from YLFW-Benchmark can match with the identities of the training data from YLFW-Dev-Train. Thus the results on YLFW-Benchmark are given under the disclaimer that they indeed do not correctly define the performance in the open-set scenario for the cases where YLFW-Dev-Train is used during training. 

\subsection{Testing on YLFW-Benchmark}
\label{exp-YLFW-Benchmark}

To investigate the properties of the proposed benchmarks and demonstrate the performance of the SOTA face recognition approaches for children's faces we stress them against the YLFW data toolset.

For simplifying the comparison we choose only the versions with the ResNet-50 \cite{Resnet2016} architecture which is trained on MS1MV2 dataset \cite{ms_celeb_face, arcface_paper}, however indeed this comparison is still not entirely correct due to the possible differences between the training settings for the models. 

We select the following set of public deep networks:
ArcFace\cite{arcface_paper}, MagFace\cite{magface}, %QualFace\cite{qualface2}, 
AdaFace\cite{AdaFace}

\begin{figure*}[htbp]
\centering
  \mbox{\subfloat[]{
            \label{fig:dev-benchs-Casia}
            \includegraphics[width=0.323\linewidth]{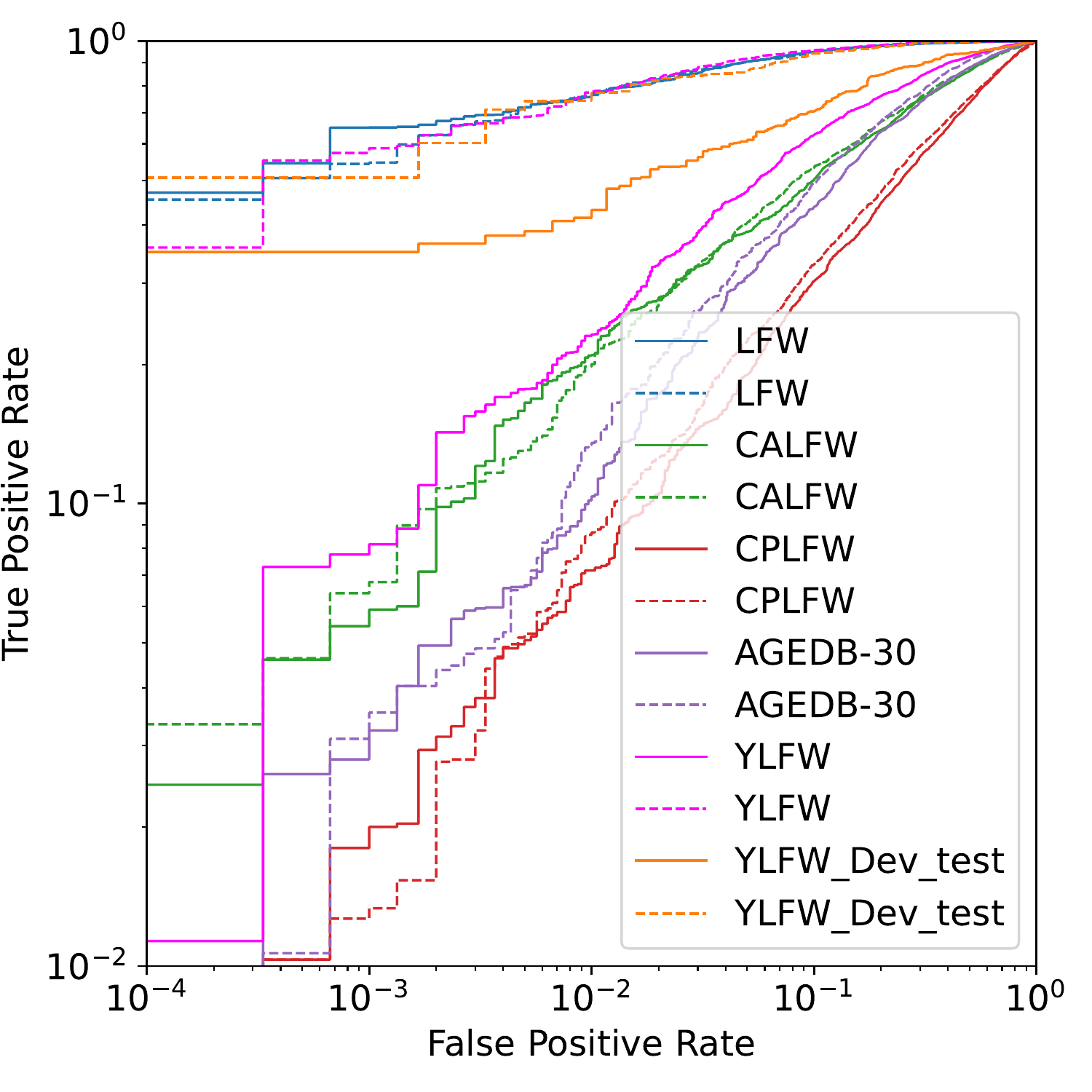}}}
  \mbox{\subfloat[]{
            \label{fig:dev-benchs-VGG} 
            \includegraphics[width=0.318\linewidth]{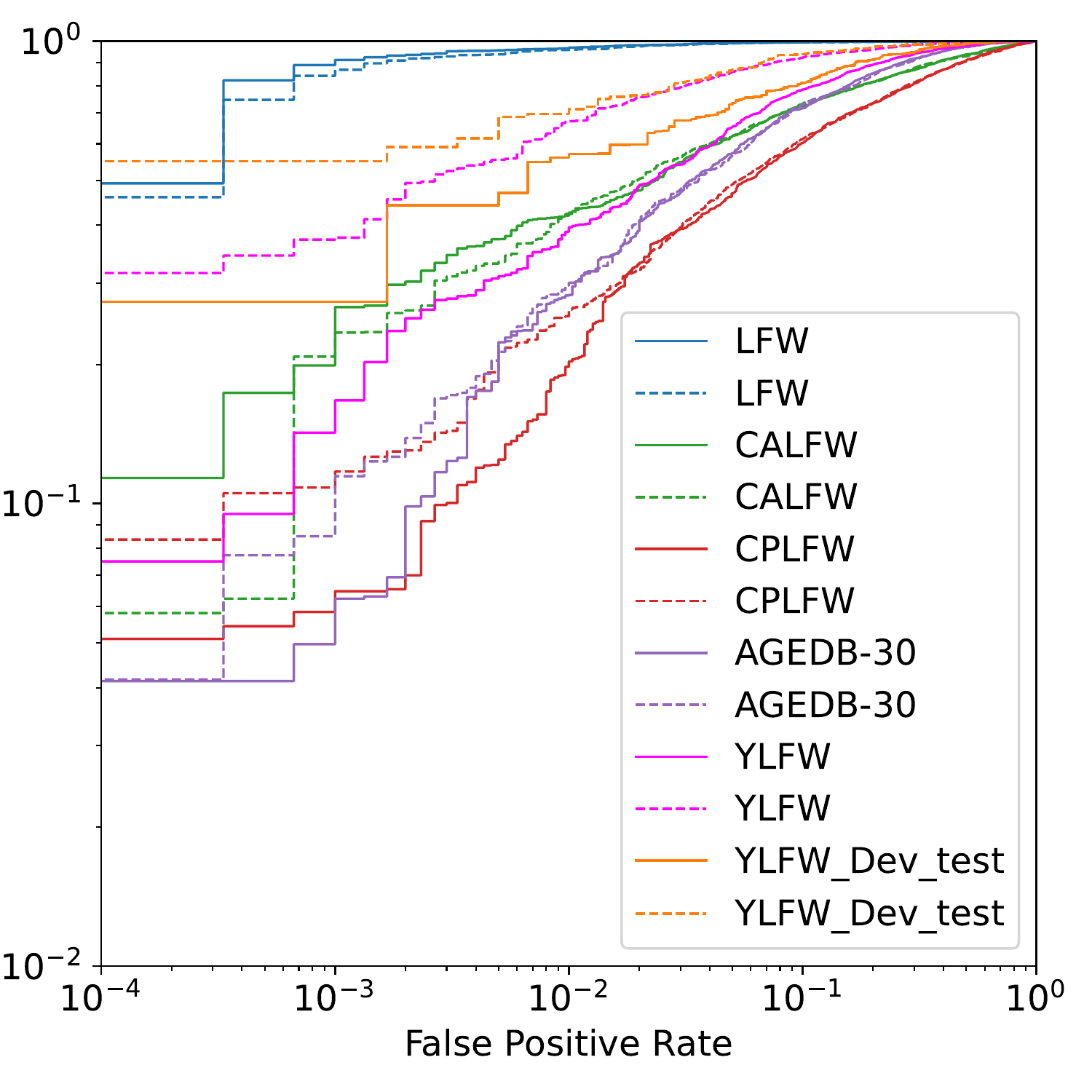}}}
  \mbox{\subfloat[]{
            \label{fig:dev-benchs-MS1MV2} 
            \includegraphics[width=0.318\linewidth]{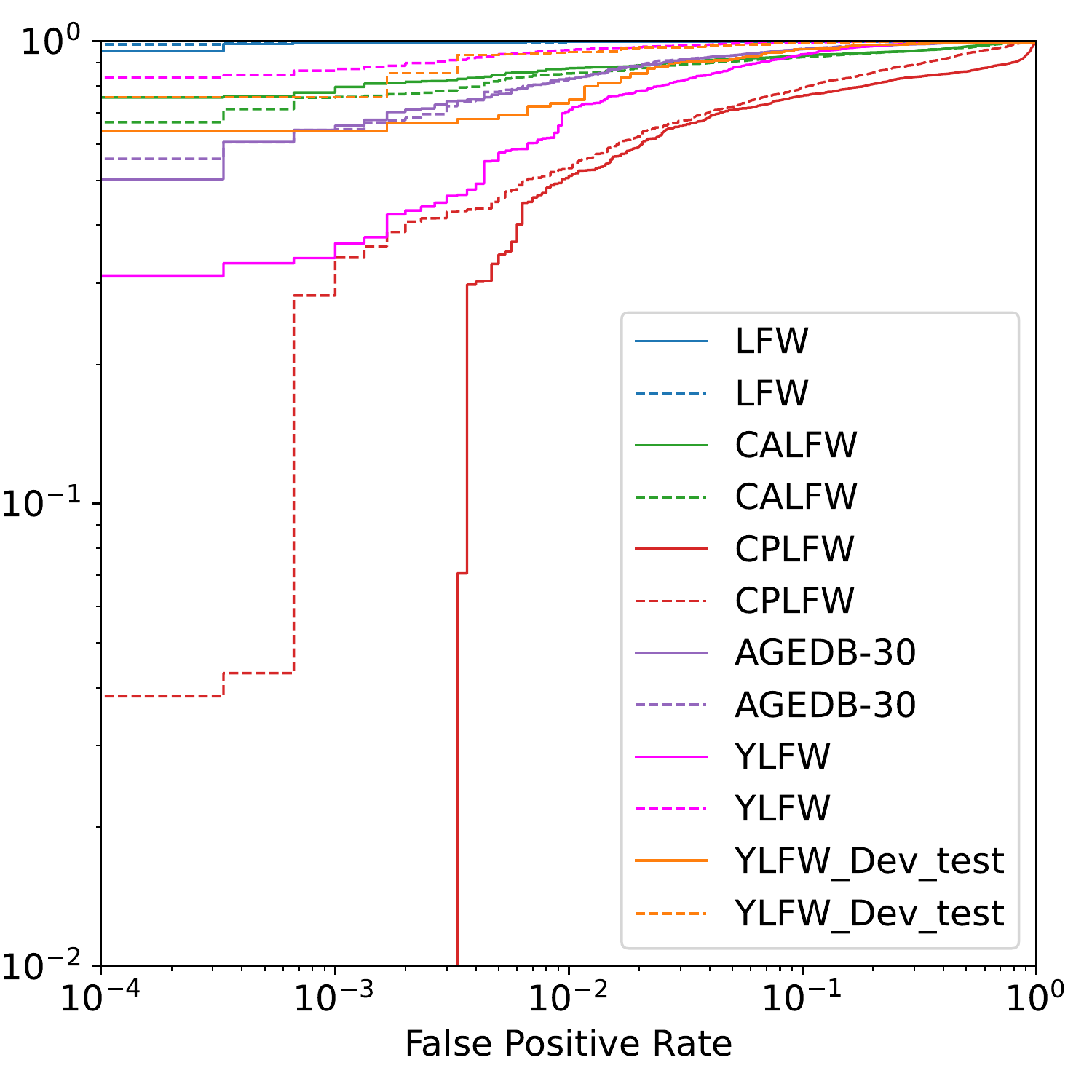}}}
\vspace*{2pt}
\caption{ROC curves of ResNet-50 trained on various baseline datasets: a - CASIA-Webface\cite{casia_webface}, b - VF2 - VGGFace2\cite{VGGface2}, c - MS1M - MS1MV2\cite{arcface_paper,ms_celeb_face}. Dashed lines correspond to the networks trained on the the baseline + YLFW-Dev-Train data. Networks are tested  on LFW, CALFW, CPLFW, AGEDB-30, YLFW-Benchmark and YLFW-Dev-Test benchmarks}
\label{fig:dev-benchs}
\end{figure*}

% Please add the following required packages to your document preamble:
% \usepackage{multirow}
\begin{table*}[htbp]
\centering
\caption{Performance metrics for LFW, CALFW, CPLFW, AGEDB-30, YLFW-Benchmark and YLFW-Dev-Test of ResNet-50 network trained on various configurations training datasets (CW - CASIA-Webface\cite{casia_webface}, VF2 - VGGFace2\cite{VGGface2}, MS1M - MS1MV2\cite{arcface_paper,ms_celeb_face} , YDTR - YLFW-Dev-Train-Balanced).}
\setlength\tabcolsep{1.5pt}
\begin{tabular}{|c|cccccccccccc|}
\hline
\multirow{3}{*}{\makecell{ \textbf{Train}\\ \textbf{Dataset} }} & \multicolumn{12}{c|}{FNMR@FMR = $\alpha$, EER, AUC of ROC}                                                                                                                                                                                                                                                   \\ \cline{2-13} 
                  & \multicolumn{4}{c|}{LFW}                                                                         & \multicolumn{4}{c|}{CALFW}                                                                         & \multicolumn{4}{c|}{CPLFW}                                                    \\ \cline{2-13} 
                  & \multicolumn{1}{c|}{$\alpha$=$10^{-1}$} & \multicolumn{1}{c|}{$\alpha$=$10^{-2}$} & \multicolumn{1}{c|}{EER} & \multicolumn{1}{c|}{AUC} & \multicolumn{1}{c|}{$\alpha$=$10^{-1}$} & \multicolumn{1}{c|}{$\alpha$=$10^{-2}$} & \multicolumn{1}{c|}{EER} & \multicolumn{1}{c|}{AUC} & \multicolumn{1}{c|}{$\alpha$=$10^{-1}$} & \multicolumn{1}{c|}{$\alpha$=$10^{-2}$} & \multicolumn{1}{c|}{EER} & AUC \\ \hline
                  CW & \multicolumn{1}{c|}{0.0520} & \multicolumn{1}{c|}{0.2314} & \multicolumn{1}{c|}{0.0721} & \multicolumn{1}{c|}{0.9798} & \multicolumn{1}{c|}{0.4981} & \multicolumn{1}{c|}{0.7902} & \multicolumn{1}{c|}{0.2753} & \multicolumn{1}{c|}{0.8010} & \multicolumn{1}{c|}{0.6970} & \multicolumn{1}{c|}{0.9283} & \multicolumn{1}{c|}{0.3721} &  0.6835\\ \hline
                  CW+YDTR & \multicolumn{1}{c|}{0.0573} & \multicolumn{1}{c|}{0.2351} & \multicolumn{1}{c|}{0.0763} & \multicolumn{1}{c|}{0.9802} & \multicolumn{1}{c|}{0.4671} & \multicolumn{1}{c|}{0.7993} & \multicolumn{1}{c|}{0.2710} & \multicolumn{1}{c|}{0.8075} & \multicolumn{1}{c|}{0.6712} & \multicolumn{1}{c|}{0.9131} & \multicolumn{1}{c|}{0.3593} &  0.6975\\ \hline
                  VF2& \multicolumn{1}{c|}{0.0056} & \multicolumn{1}{c|}{0.0316} & \multicolumn{1}{c|}{0.0193} & \multicolumn{1}{c|}{0.9975} & \multicolumn{1}{c|}{0.2756} & \multicolumn{1}{c|}{0.5740} & \multicolumn{1}{c|}{0.1896} & \multicolumn{1}{c|}{0.8893} & \multicolumn{1}{c|}{0.3976} & \multicolumn{1}{c|}{0.7969} & \multicolumn{1}{c|}{0.2376} &  0.8440\\ \hline
                  VF2+YDTR& \multicolumn{1}{c|}{0.0060} & \multicolumn{1}{c|}{0.0416} & \multicolumn{1}{c|}{0.0216} & \multicolumn{1}{c|}{0.9971} & \multicolumn{1}{c|}{0.2670} & \multicolumn{1}{c|}{0.5770} & \multicolumn{1}{c|}{0.1889} & \multicolumn{1}{c|}{0.8886} & \multicolumn{1}{c|}{0.3853} & \multicolumn{1}{c|}{0.7370} & \multicolumn{1}{c|}{0.2339} & 0.8449 \\ \hline
                  MS1M& \multicolumn{1}{c|}{0.0016} & \multicolumn{1}{c|}{0.0036} & \multicolumn{1}{c|}{0.0046} & \multicolumn{1}{c|}{0.9994} & \multicolumn{1}{c|}{0.0689} & \multicolumn{1}{c|}{0.1256} & \multicolumn{1}{c|}{0.0763} & \multicolumn{1}{c|}{0.9668} & \multicolumn{1}{c|}{0.2380} & \multicolumn{1}{c|}{0.4880} & \multicolumn{1}{c|}{0.1946} & 0.8479 \\ \hline
                  MS1M+YDTR& \multicolumn{1}{c|}{0.0010} & \multicolumn{1}{c|}{0.0040} & \multicolumn{1}{c|}{0.0050} & \multicolumn{1}{c|}{0.9996} & \multicolumn{1}{c|}{0.0763} & \multicolumn{1}{c|}{0.1480} & \multicolumn{1}{c|}{0.0816} & \multicolumn{1}{c|}{0.9634} & \multicolumn{1}{c|}{0.2053} & \multicolumn{1}{c|}{0.4683} & \multicolumn{1}{c|}{0.1636} & 0.9094 \\ \hline \hline

                & \multicolumn{4}{c|}{AGEDB-30}                                                                         & \multicolumn{4}{c|}{YLFW-Benchmark}                                                                         & \multicolumn{4}{c|}{YLFW-Dev-Test}                                                     \\ \hline
                  CW & \multicolumn{1}{c|}{0.5611} & \multicolumn{1}{c|}{0.8962} & \multicolumn{1}{c|}{0.2820} & \multicolumn{1}{c|}{0.7911} & \multicolumn{1}{c|}{0.3722} & \multicolumn{1}{c|}{0.7683} & \multicolumn{1}{c|}{0.2231} & \multicolumn{1}{c|}{0.8590} & \multicolumn{1}{c|}{0.2931} & \multicolumn{1}{c|}{0.5682} & \multicolumn{1}{c|}{0.1781} &  0.8971\\ \hline
                  CW+YDTR & \multicolumn{1}{c|}{0.5071} & \multicolumn{1}{c|}{0.8642} & \multicolumn{1}{c|}{0.2591} & \multicolumn{1}{c|}{0.8189} & \multicolumn{1}{c|}{0.0433} & \multicolumn{1}{c|}{0.2221} & \multicolumn{1}{c|}{0.0651} & \multicolumn{1}{c|}{0.9832} & \multicolumn{1}{c|}{0.0600} & \multicolumn{1}{c|}{0.2261} & \multicolumn{1}{c|}{0.0812} &  0.9779\\ \hline
                  VF2 & \multicolumn{1}{c|}{0.2830} & \multicolumn{1}{c|}{0.7173} & \multicolumn{1}{c|}{0.1736} & \multicolumn{1}{c|}{0.9072} & \multicolumn{1}{c|}{0.2136} & \multicolumn{1}{c|}{0.6040} & \multicolumn{1}{c|}{0.1506} & \multicolumn{1}{c|}{0.9256} & \multicolumn{1}{c|}{0.1883} & \multicolumn{1}{c|}{0.4300} & \multicolumn{1}{c|}{0.1383} & 0.9431\\ \hline
                  VF2+YDTR& \multicolumn{1}{c|}{0.2700} & \multicolumn{1}{c|}{0.6996} & \multicolumn{1}{c|}{0.1783} & \multicolumn{1}{c|}{0.9060} & \multicolumn{1}{c|}{0.0779} & \multicolumn{1}{c|}{0.3286} & \multicolumn{1}{c|}{0.0879} & \multicolumn{1}{c|}{0.9698} & \multicolumn{1}{c|}{0.0633} & \multicolumn{1}{c|}{0.2866} & \multicolumn{1}{c|}{0.0799} &  0.9741 \\ \hline
                  MS1M& \multicolumn{1}{c|}{0.0386} & \multicolumn{1}{c|}{0.1706} & \multicolumn{1}{c|}{0.0600} & \multicolumn{1}{c|}{0.9826} & \multicolumn{1}{c|}{0.0629} & \multicolumn{1}{c|}{0.2900} & \multicolumn{1}{c|}{0.0819} & \multicolumn{1}{c|}{0.9736} & \multicolumn{1}{c|}{0.0383} & \multicolumn{1}{c|}{0.2533} & \multicolumn{1}{c|}{0.0699} & 0.9793 \\ \hline
                  MS1M+YDTR& \multicolumn{1}{c|}{0.0373} & \multicolumn{1}{c|}{0.1680} & \multicolumn{1}{c|}{0.0606} & \multicolumn{1}{c|}{0.9841} & \multicolumn{1}{c|}{0.0070} & \multicolumn{1}{c|}{0.0420} & \multicolumn{1}{c|}{0.0243} & \multicolumn{1}{c|}{0.9973} & \multicolumn{1}{c|}{0.0083} & \multicolumn{1}{c|}{0.0533} & \multicolumn{1}{c|}{0.0300} & 0.9955\\ \hline
\end{tabular}
\label{performance_table_dev}
\end{table*}

The results for these networks are presented in Fig. \ref{fig:benchs-bench} and Table \ref{performance_table_bench}. We observe the imperfectness of the SOTA methods to the proposed in this work benchmarks. YLFW-Benchmark indeed challenges the SOTA methods on par with CALFW and CPLFW. YLFW-Dev-Test has a similar hardness to YLFW-Benchmark at high FMRs ($\sim>10^-2$), and impose easier challenge at lower FMRs.

According to Fig. \ref{fig:benchs-benchs-3}, we also observe a significant racial bias in the performance curves. However, we believe that it is also related to the negative effect of a difference in the data diversity per race in the original data for YLFW-Benchmark (see Section \ref{sec:data_collecting}).  Namely the lower data diversity of a particular race subset indeed leads to the higher similarity of the images in the match pairs.

Low values of FNMR for CALFW and CPLFW are hardly achieved since these benchmarks explicitly decrease the similarity in match pairs by the intrusion of age and pose gap (at the same time this intrusion does not increase the similarity in non-match pairs). The YLFW-Benchmark curve is more similar to LFW and does not contain that intra-class feature.
In our benchmark FNMR start increasing at lower FMR, which means that different children's identities are harder to discriminate. In a general perspective, this conforms to the intuition that children are more similar between each other rather than adults.

\subsection{Experiments on YLFW-Dev}
\label{exp-YLFW-Dev}
The YLFW-Dev can support the development of face recognition algorithms for children's faces by providing the data for both training and testing. Since the collected images do not belong to celebrities, we argue that our data can be concatenated to popular academic face datasets with low risks of introducing label conflicts.

%experiment details. backbones. image input size DONE
We perform a set of experiments to demonstrate the efficiency of such a proposal. We train a set of deep CNNs of several data configurations.  As a backbone network, we use ResNet-50 \cite{Resnet2016}, which is followed by pooling, dropout, and a dense feature layer with 512 nodes (features).  Input images (RGB 3-channel) are aligned as in \cite{arcface_paper} and resized to 112$\times$112 resolution.

%Arcface formula
As a training driver, we employ ArcFace, which allows us to learn highly discriminative feature embedding and is robust due to its simple implementation. 
ArcFace is a marginal modification of the Softmax loss, which  is usually formulated as follows:

\begin{equation}
    L_{Softmax} = \frac{1}{N}\sum_{i} -\log (\frac{e^{f_{y_i}}}{ \sum_{j}^{C} e^{f_{y_j}}}).
\end{equation}

Here the $C$ is the number of classes (identities), $N$ is a batch size, $y_i$ is the numerical index of the class of the $i-th$ sample, and $f_{y_j}$ is the $y_j-th$ element of the logits vector $\mathbf{f}$ in the last layer. 

The feature layer is usually normalized by L2 constraining the feature embeddings on a hypersphere in $\mathbb{R}^d$ space. Then $f_{y_j}$ can be represented as:  $f_{y_j} = w_j^T x_i = \cos(\theta_j)$.  
ArcFace is then obtained by adding an angular marginal penalization parameter $m$ to the positive logit:

\begin{equation}
    L_{ArcFace} = \frac{1}{N}\sum_{i} -\log (\frac{e^{s\cos(\theta_{y_i} + m)}}{e^{s\cos(\theta_{y_i} + m)} + \sum_{j\neq y_i} e^{s\cos\theta_{j}}})
\end{equation}

In our experiments in this Section, we use the ArcFace and maintain its margin $m = 0.5$, and the scaling constant $s = 30$.

%training parameters
We train the ResNet50 on three datasets: CASIA-Webface\cite{casia_webface}, VGGFace2\cite{VGGface2}, MS1MV2\cite{ms_celeb_face, arcface_paper} Then repeat the training after concatenating them with YLFW-Dev-Train-balanced. We set the following training hyperparameters: SGD optimizer with linear learning rate scheduling from 0,01 to 0,00001;  momentum 0.9;  15 epochs for VGGFAce2 and MS1MV2 and 30 epochs for CASIA-Webface; batch size 256.

%results 
The results in Fig. \ref{fig:dev-benchs} and Table \ref{performance_table_dev}  demonstrate the evident improvement in the performance in cases of augmenting the original training data with YLFW-Dev-Train-Balanced. The most significant advance is observed for YLFW-Benchmark and YLFW-Dev-Test as expected. Here the results on YLFW-Dev-Test are the better indicator since it contains the data with completely disjoint identities from the training dataset in all configurations.
%discuss in comparison with the size of the original dataset
As expected the beneficial effect of concatenating with YLFW-Dev-Test decreases with the increase of the number of classes in the original dataset, thus the most effective usage will require proper weighting of the classes from the YLFW-Dev-Train-Balanced.

\section{Conclusion}

In this work, we present a novel face data toolset, which is specifically focused on children's face recognition.  Our toolset consists of two parts YLFW-Benchmark, and YLFW-Dev for different aspects of face recognition research for the young age group. To the best of our knowledge, this data proposes the first public standardized benchmark of children's faces in the wild and the largest training dataset of children's faces.

The performed experiments demonstrate the imperfectness of modern deep face recognition approaches to children's faces.
Also, we show that the use of the presented data can facilitate accurate face recognition in the young age group.

We hope that the results of this work will stimulate and boost the research in the area of face recognition.

% \vspace{10px}
% \textbf{TODO:}

% In the further work several optional additional tasks should be completed:

% - investigate the gender properties of the data (label the data by gender)

% - investigate the age properties of the data (label the data by approximate age (Newborn (0-3 months),
% Infant (3-12 months),
% Toddler (1-5 years),
% School age (5-13 years),
% Teenager (13-18 years))

% -estimate the human accuracy on the YLFW-Benchmark dataset.

% -add QualFace to the comparison in YLFW.

% -better describe augmentations for the balanced set

\section{Acknowledgements}
 The authors would like to thank the Portuguese Mint and Official Printing Office (INCM) and the Institute of Systems and Robotics - University of Coimbra for the support of the project Facing. This work has been supported by Fundação para a Ciência e a Tecnologia (FCT) under the project UIDB/00048/2020.

%%%%%%%%% REFERENCES
{\small
\bibliographystyle{ieee_fullname}
\bibliography{egbib}
}

\end{document}